\documentclass[final,12pt]{elsarticle}

\usepackage{float}
\usepackage{array}
\usepackage[hyphens]{url}
\usepackage{hyperref}
\hypersetup{colorlinks=true,breaklinks=true}
\usepackage{threeparttable}
\usepackage{textcomp,gensymb}

\makeatletter
\def\ps@pprintTitle{%
  \let\@oddhead\@empty
  \let\@evenhead\@empty
  \def\@oddfoot{\reset@font\hfil\thepage\hfil}
  \let\@evenfoot\@oddfoot
}
\makeatother

\begin{document}

\begin{frontmatter}


\title{Explainability for Fault Detection System in Chemical Processes}

\author[label1,label3]{Georgios Gravanis} 
\ead{ggravanis@ihu.gr}

\author[label1]{Dimitrios Kyriakou}
\ead{kyriakou95@gmail.com}

\author[label3]{Spyros Voutetakis}
\ead{paris@certh.gr}

\author[label2,label3]{Simira Papadopoulou}
\ead{shmira@ihu.gr}

\author[label1]{Konstantinos Diamantaras}
\ead{kdiamant@ihu.gr}

\address[label1]{Department of Information and Electronic Engineering, International Hellenic University, Greece}
\address[label2]{Department of Industrial Engineering and Management, International Hellenic University, Greece}
\address[label3]{Chemical Process and Energy Resources Institute, Centre for Research and Technology Hellas, Greece}

\begin{abstract}

In this work, we apply and compare two state-of-the-art eXplainability Artificial Intelligence (XAI) methods, the Integrated Gradients (IG) and the SHapley Additive exPlanations (SHAP), that explain the fault diagnosis decisions of a highly accurate Long Short-Time Memory (LSTM) classifier. The classifier is trained to detect faults in a benchmark non-linear chemical process, the Tennessee Eastman Process (TEP). It is highlighted how XAI methods can help identify the subsystem of the process where the fault occurred. Using our knowledge of the process, we note that in most cases the same features are indicated as the most important for the decision, while in some cases the SHAP method seems to be more informative and closer to the root cause of the fault. Finally, since the used XAI methods are model-agnostic, the proposed approach is not limited to the specific process and can also be used in similar problems.

\end{abstract}


\end{frontmatter}

\section{Introduction}
\label{sec: Introduction}

During the so-called 4th Industrial revolution we are going through, new technologies including Augmented Reality, Big Data mining, and Deep Learning are used to optimize the operation pipelines of production facilities.
Deep Learning (DL) is highly applied with the digitization of industries and is used to enhance technologies such as Image Recognition for quality assurance (\cite{DESHPANDE20201064,ma13245755}), production variables forecasting (\cite{BRUNELLI2019248}), and Fault Detection and Diagnosis (\cite{abid2021review,dai2013model,8819956}).

However, there are still open issues concerning the adaptation of such solutions to real-life working environments. That is because the way insights are produced by such technologies is not transparent to the end users. Subsequently, the trustworthiness of the information produced by Deep Learning models is under question by the end users.

To tackle this issue, several algorithms have been developed to provide insights into Deep Learning model decisions. This relatively new research area is named eXplainable Artificial Intelligence (XAI) \cite{arrieta2020explainable}.
XAI methods mainly focus on Image classification algorithms, for example, in medical applications where images are the main diagnostic tool for several diseases \citep{van2022explainable}, and in Natural Language Process (NLP) tasks \cite{danilevsky2020survey}.

However, little research has been conducted for developing XAI methods that focus on DL models handling multivariate time series data. As this research domain is new, there are not many works published implementing explainability methods for chemical processes. Next, we present the works most related to this one.

\cite{agarwal2021explainability} propose the use of an autoencoder along with the Layerwise Relevance Propagation (LRP) algorithm to enhance the accuracy of a Fault Detection and Diagnosis (FDD) framework. The authors use a version of TEP to test the proposed architecture, with good results. 

Another approach for the explainability of a DNN in chemical processes is that of \cite{wu2021process}. The authors in their study present a Process Topology Convolutional Network (PTCN) that is based on Graph Convolutional Networks (GCN) in order to improve classification accuracy in a more transparent way.
Both works propose architectures that enhance transparency and accuracy. However, the question still remains the same: \emph{Why the classifier made its decision?} 
    
 \cite{bhakte2022explainable} uses the SHAP explainability method over a DNN on TEP but only for a few faults  i.e. IDV1, IDV2, IDV4, IDV5, IDV7, and IDV 14 (see Table \ref{tab: faults}). 
 For those faults, most FDD frameworks achieve great accuracy results since, in general, they are easily recognizable. 

With this work, we propose an approach to establish trustworthiness between the end user and the machine learning model decisions. We focus on ambiguous results and we explain the decisions of an FDD applied to a complex chemical process.  Moreover, we compare two XAI methods and we validate  the results' plausibility according to the physical interpretation of each fault. The primary motivation for this work is described by the two following Research Questions (RQ):

\begin{itemize}
    \item RQ1: Do the XAI methods results reach an agreement in the explaining variables?
    \item RQ2: Are the XAI results reasonable when applied to a chemical process time series data?

\end{itemize}

 Next, we describe how this paper is organized: Section \ref{sec: Explainability methods} briefly describes the methods used in this study to explain the deep learning algorithm decisions, whereas section \ref{sec: TEP} describes the benchmark TE process with all its variables that are used as a case study for this work. Finally, in section \ref{sec: experiments results } we present the experimental procedure and the evaluation of the results and in Section \ref{sec: Conclusions} we summarize the conclusions of this work.

\section{Explainability methods}
\label{sec: Explainability methods}

According to \cite{rojat2021explainable}, explainability methods can be separated into two main categories, namely the \emph{Ante hoc} and the \emph{Post hoc} that in general are applied either with backpropagation algorithms or with feature perturbation.
The Ante hoc explainable methods are ``embedded'' into the Neural Network model algorithms, while the Post hoc methods are applied after the training phase of the models. 

What actually differentiates the two categories is the result interpretation ability of non-experts. In the first case, when Ante hoc methods are applied to the Neural Network models, the results are mostly useful to the Machine Learning Engineer with the insights provided being valuable for algorithm optimization. 
On the contrary, Post hoc methods are more generic and aim to produce information about the decision of the Neural Network model that can be useful to a domain expert \emph{(the end-user)} that utilizes AI to solve a specific problem.  

In this work, we evaluate the explanations of two of the most popular post hoc methods namely SHAP and IG. That is because we target to build trustworthiness between the end user and the DNN model. 
Next, a short description of the two methods namely IG and SHAP will be presented.

\subsection{Integrated Gradients}
\label{sub: IG}

Integrated Gradients (IG) is a method for attributing the prediction of a neural network to its input features and it is mostly used to explain the decisions of image classifiers.
The method was first introduced by \cite{sundararajan2017axiomatic} and it belongs to the Post-hoc explainability methods. Thus, it can be applied to pre-trained ML models without any limitation on the model architecture.

According to the authors, IG ( eq. \ref{eq: IG}) complies with two basic axioms that explainability methods should fulfill, namely the \emph{Sensitivity} and the \emph{Implementation Invariance}.
\emph{Sensitivity} axiom is satisfied if an attribution method produces a non-zero attribution score when there are different predictions for every input and baseline that differ in one feature.
\emph{Implementation Invariance} axiom describes that an explainability method should produce identical attributions when two different Neural Networks have the same results.

\begin{equation}
\label{eq: IG}
IntegratedGrads_{i}(x) ::= (x_{i}-{x}'_{i}) \times \int_{a=0}^{1}\frac{\partial F(x'+a\times(x-x'))}{\partial x_{i}}da
\end{equation}

IG attributes an importance score to each feature ${x_i}$ by accumulating gradients between the current input and a baseline value.

\subsection{SHapley Additive exPlanations}
\label{sub: SHAP}

SHAP method for explaining machine learning model decisions was introduced by \cite{NIPS2017_7062}. Equation \ref{eq: shap} describes how SHAP method calculates the contribution score for each feature $i$,

\begin{equation}
\label{eq: shap}
    \phi_{i}(f,x) = \sum\limits_{z'\subseteq{x'}}\frac{|z'|!(M-|z'|-1)!}{M!}{[f_x(z')-f_x(z'\backslash{i})]}
\end{equation}
where $f$ is the model function, $x$ is the input vector, $M$ is the total number of features, $|z'|$ is the number of non-zero entries in $z'$, and $z'\subseteq{x'}$ represents all $z'$ where the non-zero entries are a subset of the non-zero entries in $x'$.

Briefly, the method estimates the contribution of each feature to the decision of any machine-learning model, by using \emph{Shapley} values from game theory. This categorizes SHAP as a post hoc explainability method because it can be implemented after the training phase of any AI model. 

\section{TEP: A base case study }
\label{sec: TEP}

\begin{figure}[htbp!]
	\centering 
	\includegraphics[width=\columnwidth]{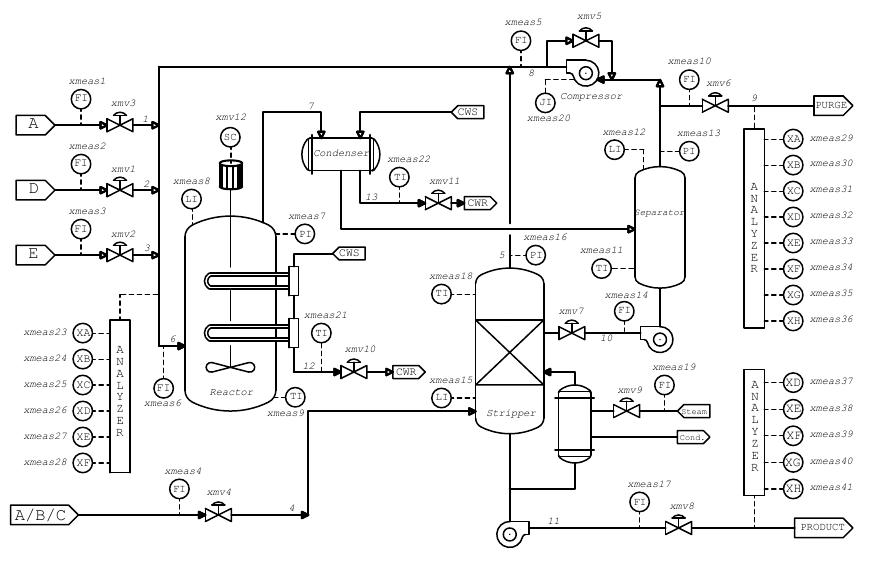}
	\caption{Tennessee Eastman Process Piping and Instrumentation Diagram.}
	\label{fig: PID_TEP}
\end{figure}

The Tennessee Eastman Process is a benchmark problem originally introduced by \cite{downsPlantwideIndustrialProcess1993}, while an updated version was introduced recently by \cite{BATHELT2015309}. Figure \ref{fig: PID_TEP} depicts the Piping and Instrumentation Diagram (P\&ID). The variables and the faults of the process are described in Table \ref{tab: variables} \& Table \ref{tab: faults}, respectively.
The initial goal of TEP was to provide a case study for the development and optimization of control methods and strategies. However, during the AI explosion era, TEP is utilized to implement frameworks for fault detection in chemical processes. (\cite{zhangDeepBeliefNetwork2017a,wu2018deep,zhang2019bidirectional}). 

This work aims to explain such DL model decisions. To achieve that, we use a highly accurate model introduced in our work \cite{GRAVANIS2022107531}, and we apply two of the most prominent XAI methods such as IG and SHAP in that model. 
Next, the experimental procedure and the results are presented.

\begin{table}[H]
	\caption{TEP variables.}
	\label{tab: variables}
    \resizebox{\columnwidth}{!}{
		\begin{threeparttable}
					\begin{tabular}{llllll}

    \hline
			Variable & Description & Units & Variable & Description & Units\\ \hline \hline
			xmeas 1 & A Feed (Stream 1)   & kSm\textsuperscript{3}/hr &  xmeas 27 & Component E  & Stream 6  \\
			xmeas 2 & D Feed (Stream 2)   & kg/hr & xmeas 28 & Component F  & Stream 6  \\
			xmeas 3 & E Feed (Stream 3)   & kg/hr & xmeas 29 & Component A  & Stream 9           \\
			xmeas4 & Total Feed (Stream 4)   &  kSm\textsuperscript{3}/hr  & xmeas 30 & Component B  & Stream 9      \\
			xmeas 5 & Recycle Flow(Stream 8)   &  kSm\textsuperscript{3}/hr  & xmeas 31 & Component C  & Stream 9       \\
			xmeas 6 & Reactor Feed Rate (Stream 6)   &  kSm\textsuperscript{3}/hr &  xmeas 32 & Component D & Stream 9 \\
			xmeas 7 & Reactor Pressure   & kPa gauge &	xmeas 33 & Component E & Stream 9    \\
			xmeas 8 & Reactor Level   & \%     & xmeas 34 & Component F  & Stream 9   \\
			xmeas 9 & Reactor Temperature   & \degree C &  xmeas 35 & Component G  & Stream 9 \\   
			xmeas 10 & Purge Rate (Stream 9)   & kscmh & xmeas 36 & Component H  & Stream 9 \\ 
			xmeas 11 & Product Sep Temp   & \degree C &   xmeas 37 & Component D  & Stream 11  \\ 
			xmeas 12 & Product Sep Level   &   \%  &  xmeas 38 & Component E  & Stream 11 \\  
			xmeas 13 & Product Sep Pressure   & kPa gauge & xmeas 39 & Component F  & Stream 11 \\
			xmeas 14 & Product Sep Underflow (Stream 10)   & m\textsuperscript{3}/hr & xmeas 40  & Component G  & Stream 11  \\ 
			xmeas 15 & Stripper Level   &  \%  & xmeas 41 & Component H  & Stream 11  \\
			xmeas 16 & Stripper Pressure   & kPa gauge &  xmv 1 & D Feed (Stream 2)   & kg/hr \\
			xmeas 17 & Stripper Underflow (Stream 11)   & m\textsuperscript{3}/hr & xmv 2 & E Feed (Stream 3)   & kg/hr \\
			xmeas 18 & Stripper Temperature   & \degree C &  xmv 3 & A Feed (Stream 1)   &   kSm\textsuperscript{3}/hr  \\
			xmeas 19 & Stripper Steam Flow   & kg/hr &  xmv 4 & Total Feed (Stream 4)   &  kSm\textsuperscript{3}/hr   \\ 
			xmeas 20 & Compressor Work   & kW  & xmv 5 & Compressor Recycle Valve   & \% \\
			xmeas 21  & Reactor Cooling Water Outlet Temp   & \degree C & xmv 6 & Purge Valve (Stream 9)   &  kSm\textsuperscript{3}/hr  \\
			xmeas 22 & Separator Cooling Water Outlet Temp   & \degree C & xmv 7 & Separator Pot Liquid Flow (Stream 10) &  m\textsuperscript{3}/hr \\
                xmeas 23 & Component A  & Stream 6 &  xmv 8  & Stripper Liquid Product Flow (Stream 11)   & m\textsuperscript{3}/hr \\ 
                xmeas 24 & Component B  & Stream 6 & xmv 9 & Stripper Steam Valve   & \%   \\
                xmeas 25 & Component C  & Stream 6 & xmv 10 & Reactor Cooling Water Flow   &  kSm\textsuperscript{3}/hr      \\
                xmeas 26 & Component D & Stream 6 & xmv 11 & Condenser Cooling Water Flow & \degree C      \\
                &  &  & xmv 12 & Agitator Speed   &  \%     \\	
                
			\hline
		\end{tabular}
			\begin{tablenotes}
				\item[1] Units of Composition measurements are mole \%
			\end{tablenotes}	
		\end{threeparttable}
	}
\end{table}

\begin{table}[H]
	\caption{Predefined disturbances in TEP.}
	\label{tab: faults}
	\resizebox{\columnwidth}{!}{
		\begin{tabular}{p{2cm}p{14cm}p{2.5cm}}
			\hline
			Disturbance & Description & Type  \\ \hline \hline
			IDV 1 & A/C feed ratio, B composition constant (Stream 4)  & step   \\
			IDV 2 & B composition, A/C ratio constant (Stream 4)  & step   \\
			IDV 3 & D feed temperature (Stream 2)  & step \\
			IDV 4 & reactor cooling water inlet temperature & step \\
			IDV 5 & condenser cooling water inlet temperature  & step \\
			IDV 6 & A feed loss (Stream 1) & step   \\ 
			IDV 7 & C header pressure loss - Reduced availability (Stream 4)  & step   \\ 
			IDV 8 & A, B, C, feed composition (Stream 4)  & random    \\
			IDV 9 & D feed temperature (Stream 2) & random    \\		
			IDV 10 & C feed temperature (Stream 4)& random     \\
			IDV 11 & Reactor cooling water inlet temperature & random     \\
			IDV 12 & Condenser cooling water inlet temperature  & random    \\ 
			IDV 13 & Reaction kinetics  & slow drift   \\ 
			IDV 14 & Reactor cooling water valve  & sticking   \\ 
			IDV 15 & Condenser cooling water valve  & sticking  \\ 
			IDV 16 & (unknown) Deviations of heat transfer within stripper (heat exchanger) & random   \\
			IDV 17  & (unknown) Deviations of heat transfer within reactor  & random   \\
			IDV 18 & (unknown) Deviations of heat transfer within condenser & random  \\
			IDV 19 & (unknown) re-cycle valve of compressor, underflow separator (stream 10), underflow stripper (stream 11) and steam valve stripper  & sticking  \\
			IDV 20  & Unknown  & random   \\ \hline
			
	\end{tabular}}
\end{table}

\section{Explainability experiments and results}
\label{sec: experiments results }

As described previously, this work applies two \emph{post hoc} explainability methods to a state-of-the-art LSTM architecture model developed for a multi-class and multivariate time series classification problem. 
The LSTM model architecture for Fault Detection and Diagnosis is described thoroughly in our previous work \cite{GRAVANIS2022107531}

To ease the evaluation process, we present the results of the explainability methods for the first 100 samples after the introduction of the disturbance in the process. 
Since process control is implemented in TEP, we can consider that measurements follow logical sequences, without large variations between neighbor ones.
With that consideration in mind, we averaged the attribution scores for each feature for a given number of input sequences. 

\begin{figure}[H]
	\centering 
	\includegraphics[trim={1cm 10cm 1cm 1cm}, width=\columnwidth]{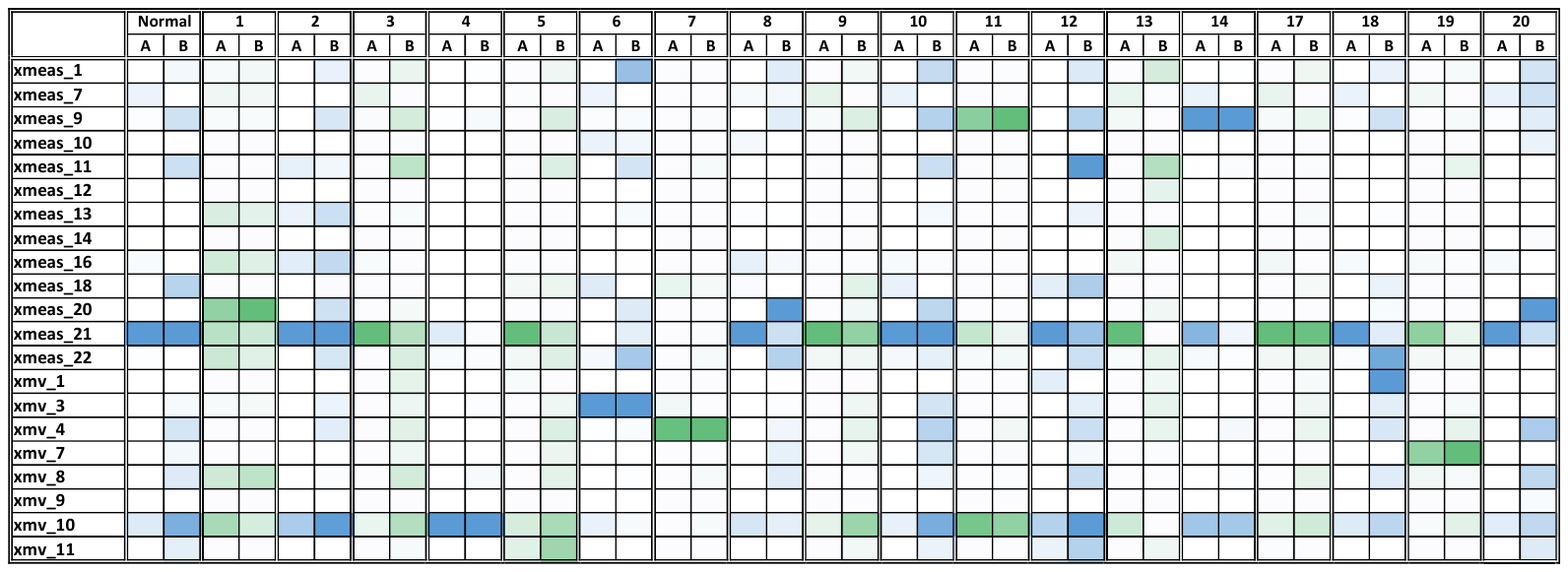}
	\caption{Heatmap with the most contributing features. The darker the color, the higher the contribution. \emph{A} columns for \emph{IG} , \emph{B} columns for \emph{SHAP}}
	\label{fig: Heatmap}
\end{figure}

In order to be able to compare the two methods, we normalized the results to identify the highly attributed features. Those features are the ones that the explainability methods indicate as important for the decision of the classifier.
The results of this process are displayed in Figure \ref{fig: Heatmap} in a qualitative way. We have to note that as proved in our previous work, not all features are important for the classifier to make its decision. To ease the reader, we excluded those features from the representation of the results.

\subsection{Result evaluation}
\label{sub: evaluation }

To evaluate the results, we grouped the faults according to the part of the process that is expected to be directly affected (Table \ref{tab: grouped disturbances}). With this grouping, we can recognize the variables that are most probable to present variation when the system is under disturbance compared to the normal operation.

\begin{table}[H]
        \footnotesize
	\caption{TEP disturbances grouped by the area affected.}
	\label{tab: grouped disturbances}
	\resizebox{\columnwidth}{!}{
\begin{tabular}{lp{10cm}ll}
\hline
\textbf{Disturbance} & Description  & \textbf{type} &\textbf{Part of the process affected} \\ \hline 
        
IDV 3  & D feed temperature (Stream 2)                     & step &           \\
IDV 4  & Reactor cooling water inlet temperature           & step &          \\
IDV 9  & D feed temperature (Stream 2)                     & random &         \\
IDV 10 & C feed temperature (Stream 4)                     & random & \textbf{Reactor (direct)}       \\
IDV 11 & Reactor cooling water inlet temperature           & random  &       \\
IDV 14 & Reactor cooling water valve                       & sticking  &     \\
IDV 17 & Deviations of heat transfer within reactor        & random  &       \\ \hline \hline
IDV 2  & B composition, A/C ratio constant (Stream 4)      & step  &    \\
IDV 8  & A, B, C, feed composition (Stream 4)              & random &        \\
IDV 13 & Reaction kinetics                                 & slow drift & \textbf{Reactor (indirect)}   \\
IDV 5  & Condenser cooling water inlet temperature         & step &          \\
IDV 12 & Condenser cooling water inlet temperature         & random &        \\
IDV 18 & Deviations of heat  transfer within condenser     & random &         \\ \hline \hline
IDV 1  & A/C feed ratio, B composition   constant (Stream 4) & step  & Stream 4 composition\\
IDV 6  & A feed loss (Stream 1)                              & step  & Stream 1  Feed \\
IDV 7  & C header pressure loss - Reduced   availability (Stream 4)  & step  & Stream 4 Feed \\
IDV 19 & Recycle valve of the compressor, underflow separator (stream 10), underflow stripper (stream 11), and steam valve stripper & sticking & Multiple areas   \\
IDV 20 & Unknown                                              & random  & unknown    \\ \hline
\end{tabular}}
\end{table}

Given that the process is under control with the strategy defined and implemented by \cite{BATHELT2015309}, it is normal for both the classifier and the explanation method to recognize important features both in measured and manipulated values.

Another important observation is that the stability of the system, which is of high complexity, is controlled mostly through the operation of its main component i.e. the reactor. 
Subsequently, for most disturbances, the implemented control strategy takes countermeasures that affect the temperature and the operation of the reactor.

\begin{table}[H]
    \footnotesize
    \caption{Reactor highly affected variables.}
    \label{tab: reactor highly affected}	
    \centering
    \begin{tabular}{ll}
    \hline
    Variable & Description  \\ \hline \hline
    xmeas 9 & Reactor temperature   \\
    xmeas 21 & Reactors' cooling water temperature \\
    xmv 10 & Reactors' cooling water valve operation \\ \hline
			
    \end{tabular}
\end{table}

That is clearly depicted in Figure \ref{fig: Heatmap} where it is shown that for faults that affect the reactor, there are three features (Table \ref{tab: reactor highly affected}) that according to XAI methods play a crucial role in classifier decisions. 

For the other faults that cannot be grouped into a specific category, the results are also reasonable. For example, when there is A feed loss (IDV 6), the methods indicate variable \emph{xmv 3} that actually controls the operation of the feeding valve for reactant A, when there is pressure loss in C header (IDV 7) they indicate  \emph{xmv 4} that is the valve for reactant C, etc.

Another important observation from the results displayed in Figure \ref{fig: Heatmap}, is that there is some differentiation between IG and SHAP methods in faults \emph{IDV 8, IDV 12, IDV 18 \& IDV 20}. After a careful examination of the results,  this differentiation indicates that the SHAP method might be more informative compared to IG.
Next, the results of two different example cases will be explained. More specifically, we will focus on the following two cases:
\begin{itemize}
\item the IDV 11 case which is a fault that directly affects the operation of the reactor heat exchanger subsystem and the XAI methods produced the same results
\item the IDV 8 case where the fault affects mostly the reaction of the process, and there is some differentiation between the methods as described before.
\end{itemize}

\subsection{IDV 11 example case}
\label{sub: IDV11 case}

\begin{figure}[H]
	\centering 
	\includegraphics[width=\columnwidth]{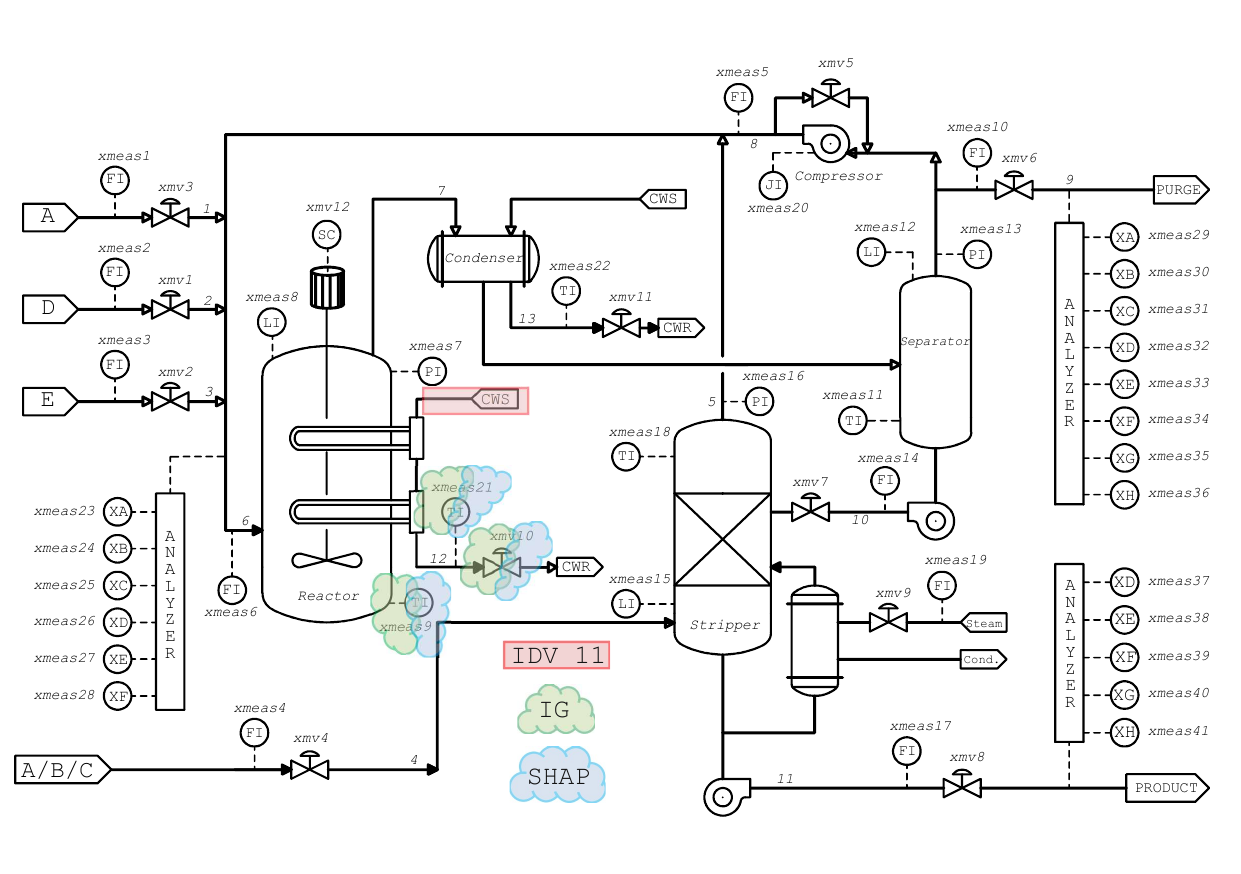}
	\caption{IG and SHAP most important features for IDV 11.}
	\label{fig: TEP fault 11}
\end{figure}

\begin{figure}

	\centering 
	\includegraphics[width=\textwidth]{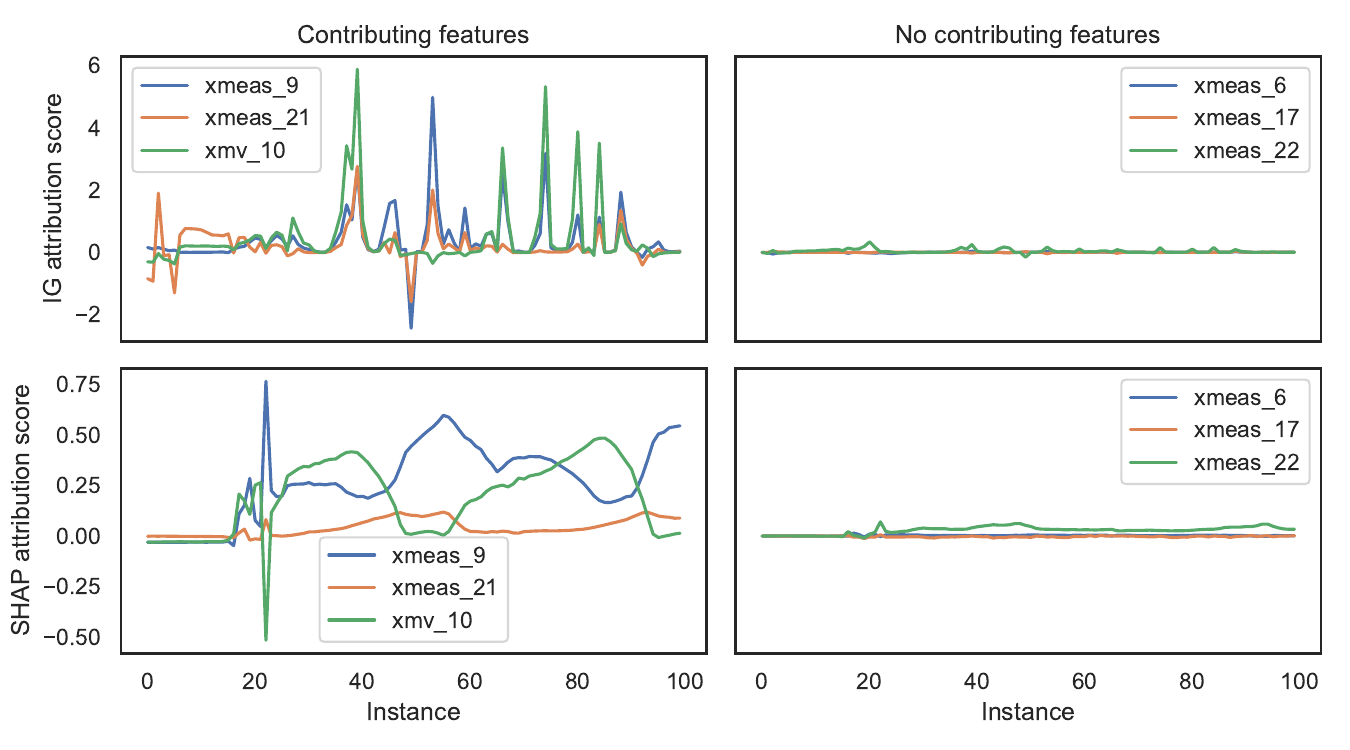}
	\caption{IG attributions for IDV 11. Left: most important features based on IG score, right: less important features based on IG score.}
	\label{fig: TEP fault 11 contributing features}
\end{figure}

\begin{figure}
	\centering 
	\includegraphics[width=\textwidth]{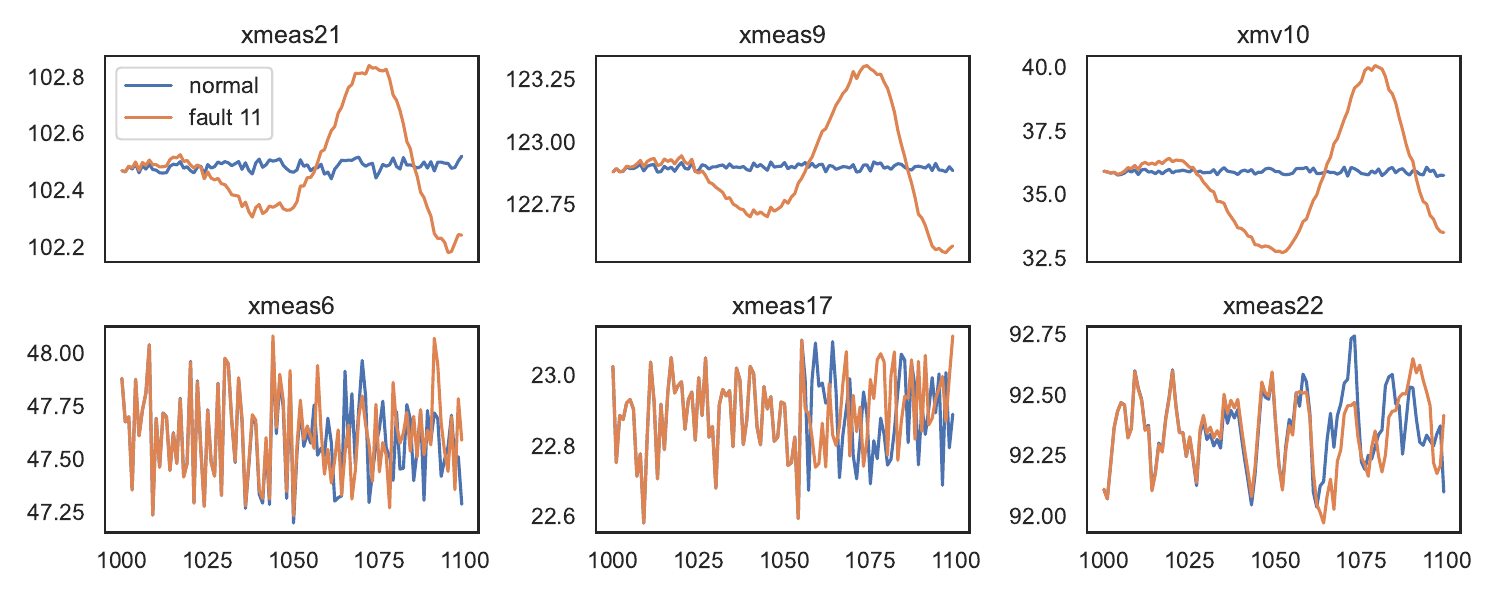}
	\caption{Most and less contributing variable behavior plots for Fault 11.}
	\label{fig: Fault 11 variables}
\end{figure}

In this case, the fault is a step variation of the cool water in the inlet of the reactor's heat exchanger (Figure \ref{fig: TEP fault 11}). The classifier recognizes the fault with 99\% accuracy, while both XAI methods indicate as most important for the classifier decisions the features \emph{xmeas 9, xmeas 21 \& xmv 10} (Table \ref{tab: Fault 11 attributions} \& Figure \ref{fig: TEP fault 11 contributing features}). 
As displayed in Figure \ref{fig: Fault 11 variables} the effect of the fault is clearly recognizable to the most important variables. 

\begin{table}
\caption{Fault 11 IG and SHAP attribution scores after normalization.}
\label{tab: Fault 11 attributions}
\centering
\resizebox{\columnwidth}{!}{

\begin{tabular}{lrr|lrr|lrr|lrr}
\hline
feature           & \multicolumn{1}{l}{IG} & \multicolumn{1}{l|}{SHAP} & feature            & \multicolumn{1}{l}{IG} & \multicolumn{1}{l|}{SHAP} & feature   & \multicolumn{1}{l}{IG} & \multicolumn{1}{l|}{SHAP} & feature          & \multicolumn{1}{l}{IG} & \multicolumn{1}{l}{SHAP} \\ \hline
xmeas\_1          & -0.48                  & 0.08                      & xmeas\_15          & -0.21                  & -0.24                     & xmeas\_29 & -0.20                  & -0.29                     & xmv\_1           & -0.40                  & -0.32                    \\
xmeas\_2          & -0.19                  & -0.32                     & xmeas\_16          & -0.08                  & -0.11                     & xmeas\_30 & -0.36                  & -0.19                     & xmv\_2           & -0.22                  & -0.23                    \\
xmeas\_3          & -0.25                  & -0.33                     & xmeas\_17          & -0.20                  & -0.37                     & xmeas\_31 & -0.19                  & -0.30                     & xmv\_3           & -0.74                  & 0.01                     \\
xmeas\_4          & -0.16                  & -0.16                     & xmeas\_18          & 0.01                   & -0.10                     & xmeas\_32 & -0.18                  & -0.32                     & xmv\_4           & -1.24                  & 0.39                     \\
xmeas\_5          & -0.16                  & -0.27                     & xmeas\_19          & -0.26                  & -0.33                     & xmeas\_33 & -0.15                  & -0.31                     & xmv\_5           & -0.20                  & -0.32                    \\
xmeas\_6          & -0.21                  & -0.25                     & xmeas\_20          & -0.15                  & -0.08                     & xmeas\_34 & -0.28                  & -0.14                     & xmv\_6           & -0.17                  & -0.15                    \\
xmeas\_7          & 0.01                   & -0.29                     & \textbf{xmeas\_21} & \textbf{2.11}          & \textbf{0.65}             & xmeas\_35 & -0.20                  & -0.38                     & xmv\_7           & -0.14                  & 0.09                     \\
xmeas\_8          & -0.16                  & -0.33                     & xmeas\_22          & 0.23                   & 0.35                      & xmeas\_36 & -0.23                  & -0.34                     & xmv\_8           & -0.18                  & 0.09                     \\
\textbf{xmeas\_9} & \textbf{4.34}          & \textbf{5.74}             & xmeas\_23          & -0.18                  & -0.34                     & xmeas\_37 & -0.21                  & -0.33                     & xmv\_9           & -0.20                  & -0.32                    \\
xmeas\_10         & -0.18                  & -0.19                     & xmeas\_24          & -0.20                  & -0.23                     & xmeas\_38 & -0.22                  & -0.30                     & \textbf{xmv\_10} & \textbf{5.01}          & \textbf{4.00}            \\
xmeas\_11         & -0.30                  & 0.11                      & xmeas\_25          & -0.16                  & -0.33                     & xmeas\_39 & -0.20                  & -0.29                     & xmv\_11          & -0.26                  & -0.20                    \\
xmeas\_12         & -0.19                  & -0.32                     & xmeas\_26          & -0.15                  & -0.31                     & xmeas\_40 & -0.20                  & -0.34                     & xmv\_12          & -0.20                  & -0.32                    \\
xmeas\_13         & -0.53                  & -0.12                     & xmeas\_27          & -0.20                  & -0.29                     & xmeas\_41 & -0.14                  & -0.28                     &                  &                        &                          \\
xmeas\_14         & -0.18                  & -0.31                     & xmeas\_28          & -0.24                  & -0.23                     &           &                        &                           &                  &                        &                          \\ \hline
\end{tabular}}
\end{table}

\subsection{IDV 8 example case}
\label{sub: IDV 8}

\begin{figure}[H]
	\centering 
	\includegraphics[width=\columnwidth]{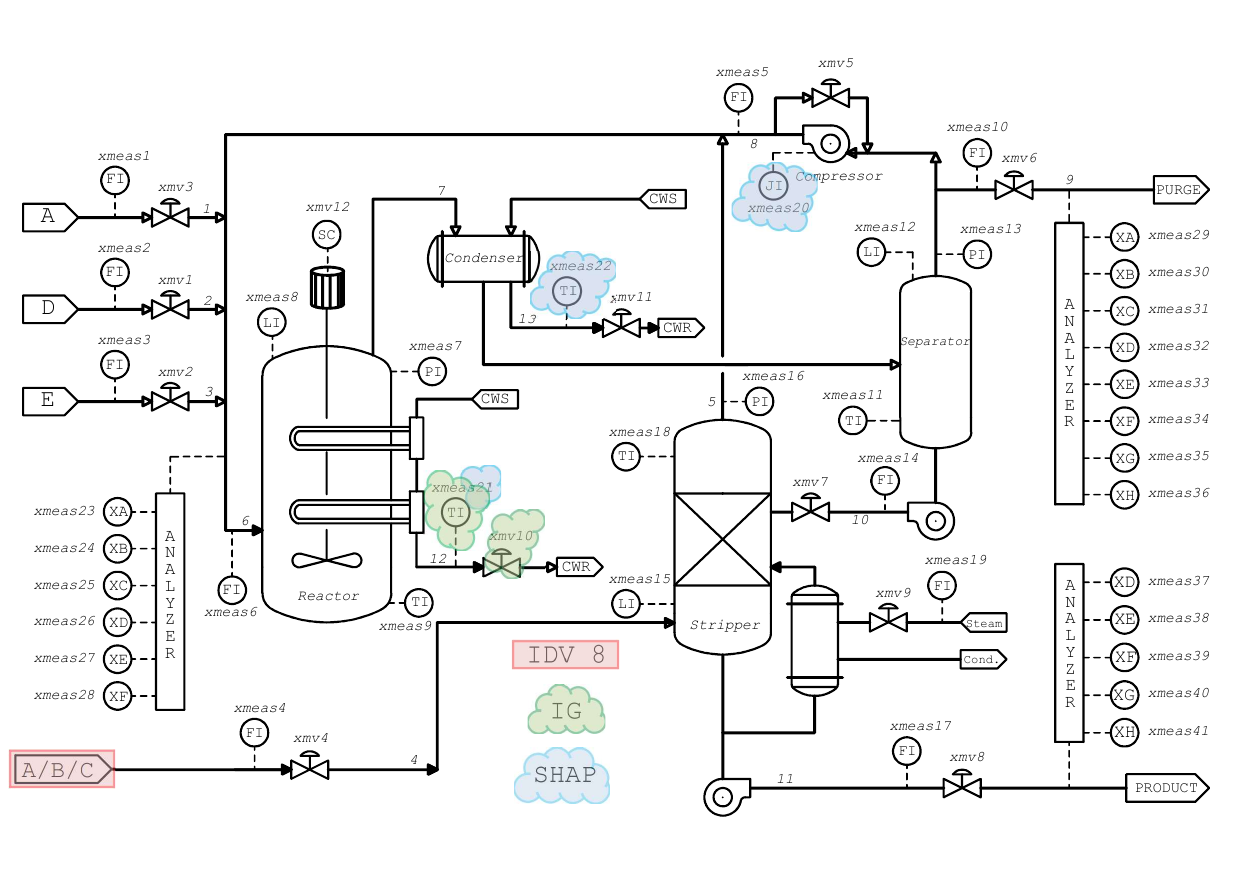}
	\caption{IG and SHAP most important features for IDV 8}
	\label{fig: TEP fault 8}
\end{figure}

\begin{figure}[H]
	\centering 
	\includegraphics[width=\textwidth]{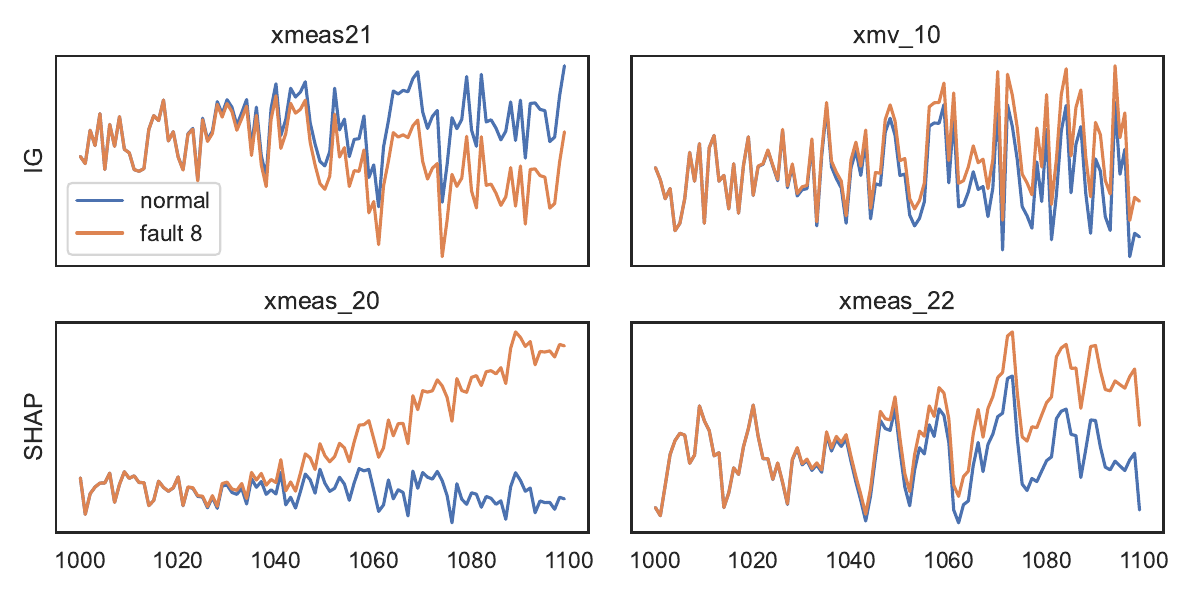}
	\caption{IDV 8 most important variables according to IG and SHAP }
	\label{fig: variables_8}
\end{figure}

\begin{figure}[H]
	\centering 
	\includegraphics[scale=0.8]{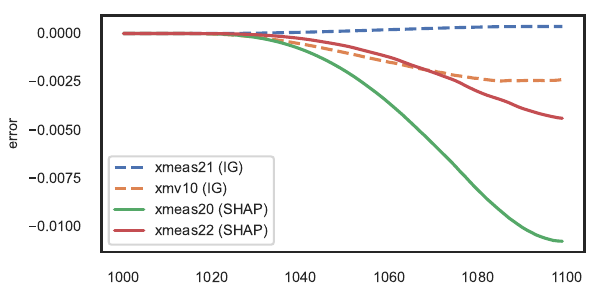}
	\caption{IDV 8 most important variables according to IG and SHAP }
	\label{fig: variables_8_4}
\end{figure}

In the case of IDV 8 (Figure \ref{fig: TEP fault 8}), the fault is a random variation in the composition of the reactants of stream 4, and the classifier achieves 99\% accuracy. Here there is some differentiation between the methods since SHAP indicates \emph{xmeas 20 \& xmeas 22} as the most important features while for IG the most attributed features from IG are \emph{xmeas 21 \& xmv 10}. As shown in  Figure \ref{fig: variables_8}  all four variables have a relatively large differentiation from the normal operation. However, the variables that the SHAP method indicates as the most important, have greater error variation compared to the variables indicated by the IG method (Figure \ref{fig: variables_8_4}). 
In general, the error is expected to be greater for variables belonging to subsystems adjacent to the root cause, while the error of variables which are further away is absorbed by the control system of the process.
This could be an indication that the SHAP method in some cases may be more informative than the IG.

\begin{table}
\caption{Fault 8 IG and SHAP attribution scores after normalization.}
\label{tab: Fault 8 attributions}
\centering
\resizebox{\columnwidth}{!}{

\begin{tabular}{lrr|lrr|lrr|lrr}
\hline
feature           & \multicolumn{1}{l}{IG} & \multicolumn{1}{l|}{SHAP} & feature            & \multicolumn{1}{l}{IG} & \multicolumn{1}{l|}{SHAP} & feature   & \multicolumn{1}{l}{IG} & \multicolumn{1}{l|}{SHAP} & feature          & \multicolumn{1}{l}{IG} & \multicolumn{1}{l}{SHAP} \\ \hline
xmeas\_1          & -0.27                  & 0.98                      & xmeas\_15          & -0.37                  & -0.22                     & xmeas\_29 & 0.80                  & -0.03                     & xmv\_1           & -0.60                  & -0.77                    \\
xmeas\_2          & -0.20                  & -0.39                     & xmeas\_16          & 0.93                  & 0.33                     & xmeas\_30 & -0.31                  & -0.31                     & xmv\_2           & -0.34                  & -0.32                    \\
xmeas\_3          & -0.28                  & -0.34                     & xmeas\_17          & -0.29                  & -0.40                     & xmeas\_31 & 0.57                  & -0.18                     & xmv\_3           & -0.79                  & 0.15                     \\
xmeas\_4          & -0.26                  & -0.36                     & xmeas\_18          & 0.27                   & -3.19                     & xmeas\_32 & -0.19                  & -0.59                     & xmv\_4           & -1.40                  & 0.48                     \\
xmeas\_5          & -0.20                  & 0.18                     & xmeas\_19          & -0.18                  & -0.43                     & xmeas\_33 & -0.15                  & -0.37                     & xmv\_5           & -0.18                  & -0.41                    \\
xmeas\_6          & -0.31                  & 0.32                     & xmeas\_20          & -0.18                  & \textbf{5.04}                     & xmeas\_34 & -0.40                  & -0.06                     & xmv\_6           & 0.38                  & -0.22                    \\
xmeas\_7          & 0.38                   & 0.38                     & xmeas\_21       & \textbf{6.37}                   & \textbf{1.60}                      & xmeas\_35 & -0.22                  & -0.37                     & xmv\_7           & -0.29                  & 0.74                     \\
xmeas\_8          & 0.03                  & -0.35                     & xmeas\_22          & -0.24                   & \textbf{2.28}                      & xmeas\_36 & -0.33                  & -0.49                     & xmv\_8           & -0.50                  & 0.98                     \\
xmeas\_9 & -0.39          & 0.93             & xmeas\_23          & 0.39                  & -0.13                     
        & xmeas\_37 & -0.12                  & -0.39                     & xmv\_9           & -0.18                  & -0.41                    \\
xmeas\_10         & 0.44                  & -0.37                     & xmeas\_24          & -0.18                  & -0.27                     & xmeas\_38 & -0.20                  & -0.17                     & \textbf{xmv\_10} & \textbf{1.62}          & 0.85            \\
xmeas\_11         & -1.31                  & -0.12                      & xmeas\_25          & 0.43                  & 0.09                     & xmeas\_39 & -0.17                  & -0.47                     & xmv\_11          & -0.25                  & -0.53                    \\
xmeas\_12         & -0.12                  & -0.37                     & xmeas\_26          & -0.10                  & -0.37                     & xmeas\_40 & -0.19                  & -0.40                     & xmv\_12          & -0.18                  & -0.41                    \\
xmeas\_13         & 0.16                  & 0.04                     & xmeas\_27          & -0.14                  & -0.18                     & xmeas\_41 & -0.16                  & -0.42                     &                  &                        &                          \\
xmeas\_14         & -0.23                  & -0.39                     & xmeas\_28          & -0.36                  & -0.16                     &           &                        &                           &                  &                        &                          \\ \hline
\end{tabular}}
\end{table}

\section{Conclusions}
\label{sec: Conclusions}

In this work two state-of-the-art methods namely IG and SHAP are used to explain the decisions of a highly accurate LSTM model trained to identify faults in a non-linear chemical process. The results of both methods are reasonable and in most cases, they agree in the list with the same features as the most important. In some cases i.e. \emph{IDV 8, IDV 12 \& IDV 18}, the SHAP method seems to be more informative than IG. However, both IG and SHAP showed efficacy and consistency in producing valuable insights and in enlightening any obscure points on the decisions of Deep Learning models in a chemical process application. Finally, since the proposed approach uses Post - hoc XAI methods, it can be adapted and used in a large variety of industrial and chemical process applications.

\bibliographystyle{model5-names}
\biboptions{authoryear}

\bibliography{bibliography}

@inproceedings{sundararajan2017axiomatic,
  title={Axiomatic attribution for deep networks},
  author={Sundararajan, Mukund and Taly, Ankur and Yan, Qiqi},
  booktitle={International conference on machine learning},
  pages={3319--3328},
  year={2017},
  organization={PMLR}
}

@article{GRAVANIS2022107531,
title = {Fault detection and diagnosis for non-linear processes empowered by dynamic neural networks},
journal = {Computers \& Chemical Engineering},
volume = {156},
pages = {107531},
year = {2022},
issn = {0098-1354},
doi = {https://doi.org/10.1016/j.compchemeng.2021.107531},
url = {https://www.sciencedirect.com/science/article/pii/S0098135421003094},
author = {Georgios Gravanis and Ioannis Dragogias and Konstantinos Papakiriakos and Chrysovalantou Ziogou and Konstantinos Diamantaras},
}

@article{downsPlantwideIndustrialProcess1993,
	series = {Industrial Challenge Problems in Process Control},
	title = {A Plant-Wide Industrial Process Control Problem},
	volume = {17},
	issn = {0098-1354},
	number = {3},
	journal = {Computers \& Chemical Engineering},
	doi = {10.1016/0098-1354(93)80018-I},
	author = {Downs, J. J. and Vogel, E. F.},
	month = mar,
	year = {1993},
	pages = {245-255},
}

@article{BATHELT2015309,
title = "Revision of the Tennessee Eastman Process Model",
journal = "IFAC-PapersOnLine",
volume = "48",
number = "8",
pages = "309 - 314",
year = "2015",
note = "9th IFAC Symposium on Advanced Control of Chemical Processes ADCHEM 2015",
issn = "2405-8963",
doi = "https://doi.org/10.1016/j.ifacol.2015.08.199",
url = "http://www.sciencedirect.com/science/article/pii/S2405896315010666",
author = "Andreas Bathelt and N. Lawrence Ricker and Mohieddine Jelali"
}

@article{rojat2021explainable,
  title={Explainable artificial intelligence (xai) on timeseries data: A survey},
  author={Rojat, Thomas and Puget, Rapha{\"e}l and Filliat, David and Del Ser, Javier and Gelin, Rodolphe and D{\'\i}az-Rodr{\'\i}guez, Natalia},
  journal={arXiv preprint arXiv:2104.00950},
  year={2021}
}

@incollection{NIPS2017_7062,
title = {A Unified Approach to Interpreting Model Predictions},
author = {Lundberg, Scott M and Lee, Su-In},
booktitle = {Advances in Neural Information Processing Systems 30},
editor = {I. Guyon and U. V. Luxburg and S. Bengio and H. Wallach and R. Fergus and S. Vishwanathan and R. Garnett},
pages = {4765--4774},
year = {2017},
publisher = {Curran Associates, Inc.},
url = {http://papers.nips.cc/paper/7062-a-unified-approach-to-interpreting-model-predictions.pdf}
}

@article{agarwal2021explainability,
  title={Explainability: Relevance based dynamic deep learning algorithm for fault detection and diagnosis in chemical processes},
  author={Agarwal, Piyush and Tamer, Melih and Budman, Hector},
  journal={Computers \& Chemical Engineering},
  volume={154},
  pages={107467},
  year={2021},
  publisher={Elsevier}
}

@article{wu2021process,
  title={Process topology convolutional network model for chemical process fault diagnosis},
  author={Wu, Deyang and Zhao, Jinsong},
  journal={Process Safety and Environmental Protection},
  volume={150},
  pages={93--109},
  year={2021},
  publisher={Elsevier}
}

@article{bhakte2022explainable,
  title={An explainable artificial intelligence based approach for interpretation of fault classification results from deep neural networks},
  author={Bhakte, Abhijit and Pakkiriswamy, Venkatesh and Srinivasan, Rajagopalan},
  journal={Chemical Engineering Science},
  volume={250},
  pages={117373},
  year={2022},
  publisher={Elsevier}
}

@article{arrieta2020explainable,
  title={Explainable Artificial Intelligence (XAI): Concepts, taxonomies, opportunities and challenges toward responsible AI},
  author={Arrieta, Alejandro Barredo and D{\'\i}az-Rodr{\'\i}guez, Natalia and Del Ser, Javier and Bennetot, Adrien and Tabik, Siham and Barbado, Alberto and Garc{\'\i}a, Salvador and Gil-L{\'o}pez, Sergio and Molina, Daniel and Benjamins, Richard and others},
  journal={Information fusion},
  volume={58},
  pages={82--115},
  year={2020},
  publisher={Elsevier}
}

@article{van2022explainable,
  title={Explainable artificial intelligence (XAI) in deep learning-based medical image analysis},
  author={van der Velden, Bas HM and Kuijf, Hugo J and Gilhuijs, Kenneth GA and Viergever, Max A},
  journal={Medical Image Analysis},
  pages={102470},
  year={2022},
  publisher={Elsevier}
}

@Article{ma13245755,
AUTHOR = {Yang, Jing and Li, Shaobo and Wang, Zheng and Dong, Hao and Wang, Jun and Tang, Shihao},
TITLE = {Using Deep Learning to Detect Defects in Manufacturing: A Comprehensive Survey and Current Challenges},
JOURNAL = {Materials},
VOLUME = {13},
YEAR = {2020},
NUMBER = {24},
ARTICLE-NUMBER = {5755},
URL = {https://www.mdpi.com/1996-1944/13/24/5755},
PubMedID = {33339413},
ISSN = {1996-1944},
DOI = {10.3390/ma13245755}
}

@article{DESHPANDE20201064,
title = {One-Shot Recognition of Manufacturing Defects in Steel Surfaces},
journal = {Procedia Manufacturing},
volume = {48},
pages = {1064-1071},
year = {2020},
note = {48th SME North American Manufacturing Research Conference, NAMRC 48},
issn = {2351-9789},
doi = {https://doi.org/10.1016/j.promfg.2020.05.146},
url = {https://www.sciencedirect.com/science/article/pii/S2351978920315985},
author = {Aditya M. Deshpande and Ali A. Minai and Manish Kumar},
}

@article{BRUNELLI2019248,
title = {Deep Learning-based Production Forecasting in Manufacturing: a Packaging Equipment Case Study},
journal = {Procedia Manufacturing},
volume = {38},
pages = {248-255},
year = {2019},
note = {29th International Conference on Flexible Automation and Intelligent Manufacturing ( FAIM 2019), June 24-28, 2019, Limerick, Ireland, Beyond Industry 4.0: Industrial Advances, Engineering Education and Intelligent Manufacturing},
issn = {2351-9789},
doi = {https://doi.org/10.1016/j.promfg.2020.01.033},
url = {https://www.sciencedirect.com/science/article/pii/S2351978920300342},
author = {Luca Brunelli and Chiara Masiero and Diego Tosato and Alessandro Beghi and Gian Antonio Susto},
}

@ARTICLE{8819956,  author={Saufi, Syahril Ramadhan and Ahmad, Zair Asrar Bin and Leong, Mohd Salman and Lim, Meng Hee},  journal={IEEE Access},   title={Challenges and Opportunities of Deep Learning Models for Machinery Fault Detection and Diagnosis: A Review},   year={2019},  volume={7},  number={},  pages={122644-122662},  doi={10.1109/ACCESS.2019.2938227}}

@article{abid2021review,
  title={A review on fault detection and diagnosis techniques: basics and beyond},
  author={Abid, Anam and Khan, Muhammad Tahir and Iqbal, Javaid},
  journal={Artificial Intelligence Review},
  volume={54},
  number={5},
  pages={3639--3664},
  year={2021},
  publisher={Springer}
}

@article{dai2013model,
  title={From model, signal to knowledge: A data-driven perspective of fault detection and diagnosis},
  author={Dai, Xuewu and Gao, Zhiwei},
  journal={IEEE Transactions on Industrial Informatics},
  volume={9},
  number={4},
  pages={2226--2238},
  year={2013},
  publisher={IEEE}
}

@article{danilevsky2020survey,
  title={A survey of the state of explainable AI for natural language processing},
  author={Danilevsky, Marina and Qian, Kun and Aharonov, Ranit and Katsis, Yannis and Kawas, Ban and Sen, Prithviraj},
  journal={arXiv preprint arXiv:2010.00711},
  year={2020}
}

@article{zhangDeepBeliefNetwork2017a,
	series = {In Honor of {{Professor Rafiqul Gani}}},
	title = {A Deep Belief Network Based Fault Diagnosis Model for Complex Chemical Processes},
	volume = {107},
	issn = {0098-1354},
	journal = {Computers \& Chemical Engineering},
	doi = {10.1016/j.compchemeng.2017.02.041},
	author = {Zhang, Zhanpeng and Zhao, Jinsong},
	month = dec,
	year = {2017},
	pages = {395-407},
}

@article{wu2018deep,
  title={Deep convolutional neural network model based chemical process fault diagnosis},
  author={Wu, Hao and Zhao, Jinsong},
  journal={Computers \& Chemical Engineering},
  volume={115},
  pages={185--197},
  year={2018},
  publisher={Elsevier}
}

@article{zhang2019bidirectional,
  title={Bidirectional recurrent neural network-based chemical process fault diagnosis},
  author={Zhang, Shuyuan and Bi, Kexin and Qiu, Tong},
  journal={Industrial \& Engineering Chemistry Research},
  volume={59},
  number={2},
  pages={824--834},
  year={2019},
  publisher={ACS Publications}
}

\end{document}